\documentclass[letterpaper, 10pt, conference]{IEEEtran}
\IEEEoverridecommandlockouts

\usepackage[T1]{fontenc}

\usepackage{siunitx}
\usepackage{cite}
\usepackage{amsmath,amssymb,amsfonts}
\usepackage{algorithmicx}
\usepackage{graphicx}
\usepackage{textcomp}
\usepackage{xcolor}
\usepackage{algorithm}
\usepackage{algorithmicx}
\usepackage[noend]{algpseudocode}
\usepackage{mathtools}
\usepackage{multirow}
\usepackage{multicol}
\usepackage{subfig}

\colorlet{mygray}{black!30}
\colorlet{mygreen}{green!60!blue}
\colorlet{mymauve}{red!60!blue}

\def\BibTeX{{\rm B\kern-.05em{\sc i\kern-.025em b}\kern-.08em
    T\kern-.1667em\lower.7ex\hbox{E}\kern-.125emX}}
    
\makeatletter
\newcommand\fs@spaceruled{\def\@fs@cfont{\bfseries}\let\@fs@capt\floatc@ruled
  \def\@fs@pre{\vspace{0.7\baselineskip}\hrule height.8pt depth0pt \kern2pt}%
  \def\@fs@post{\kern2pt\hrule\relax}%
  \def\@fs@mid{\kern2pt\hrule\kern2pt}%
  \let\@fs@iftopcapt\iftrue}
\makeatother

\usepackage{nicefrac}       
\usepackage{microtype}      
\usepackage{diagbox}
\usepackage{tikz}
\usepackage{adjustbox}
\usepackage{amsthm}

\usepackage{flushend}

\usepackage{listings}
\newcommand{\libname}{OpenMORE}

\newcommand{\fillbox}[3]
{\bgroup
  \dimen1=#1\relax
  \dimen2=#2\relax
  \sbox0{\includegraphics[width=#1]{#3}}%
  \ifdim\ht0>\dimen2
    \dimen0=\dimexpr \ht0-\dimen2\relax
    \adjustbox{clip=true,trim=0pt 0.5\dimen0 0pt 0.5\dimen0}{\usebox0}%
  \else
    \sbox0{\includegraphics[height=#2]{#3}}%
    \ifdim\wd0>\dimen1
      \dimen0=\dimexpr \wd0-\dimen1\relax
      \adjustbox{clip=true,trim=0.5\dimen0 0pt 0.5\dimen0 0pt}{\usebox0}%
    \else
      \usebox0
    \fi
  \fi
\egroup}

\newcommand{\DocTitle}{{\libname}: an open-source tool for sampling-based path replanning in ROS}

\newcommand{\TeaserImage}
{
\begin{center}
    \centering
    \setcounter{figure}{0}
    \captionsetup{type=figure}
    \subfloat[][Replanning in human-robot collaboration]
    {
     \includegraphics[width=5.3cm,height=3.6cm]{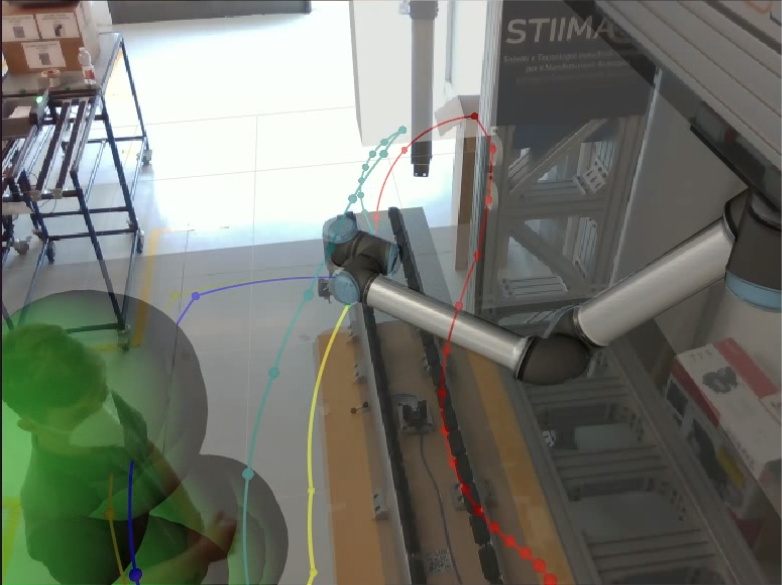}
      \label{subfig: hrc}
    }\,
    \subfloat[][Simulated manipulator]
    {
     \includegraphics[width=5.3cm,height=3.6cm]{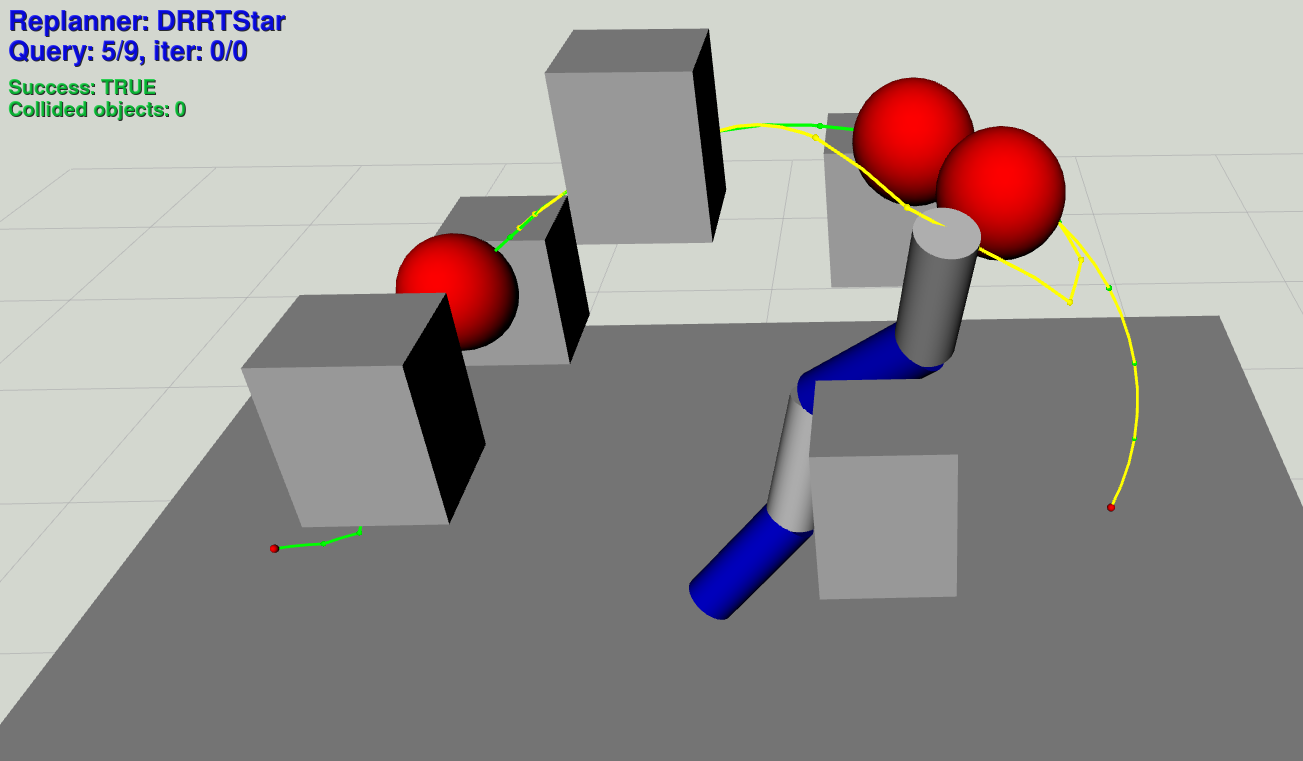}
     \label{subfig: simulation arm}
    }\,
    \subfloat[][Simulated 3D robot]
    {
     \includegraphics[width=5.3cm,height=3.6cm]{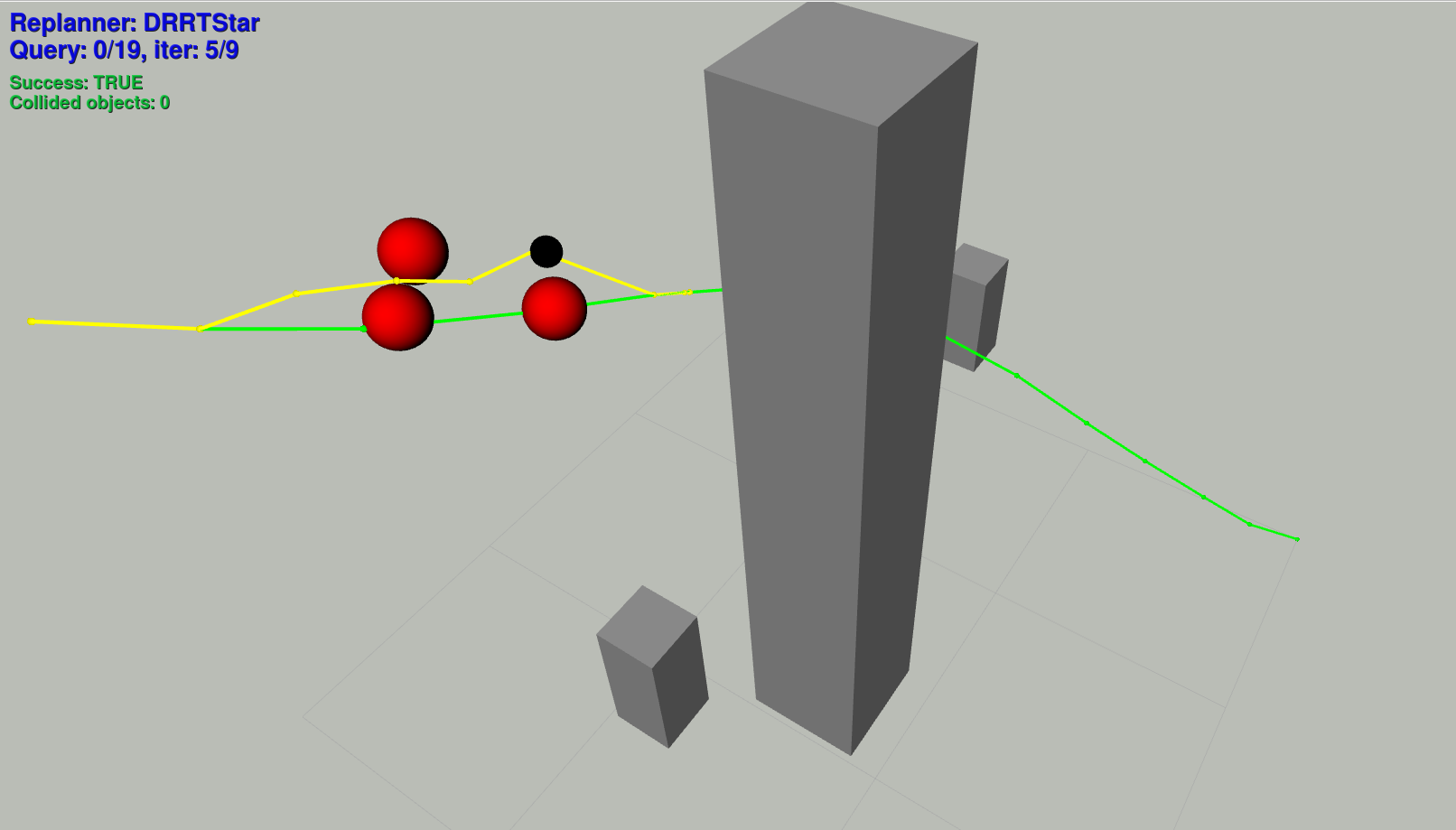}
     \label{subfig: simulation point}
    }
    \captionof{figure}{Examples of application of {\libname}. In simulated environments, red spheres are unexpected obstacles, green and yellow lines are the initial and the current path, respectively.}
    \label{fig:examples}
\end{center}%
\vspace{-0.2cm}
}

\makeatletter
\apptocmd{\@maketitle}{\centering\TeaserImage}{}{}
\makeatother

\begin{document}

\title{\DocTitle}

\author{
Cesare Tonola$^{1,2}$, Manuel Beschi$^{1,2}$, Marco Faroni$^{3}$, Nicola Pedrocchi$^{1}$  
\thanks{$^{1}$ STIIMA-CNR - Institute of Intelligent Industrial Technologies and Systems, National Research Council of Italy 
{\tt\small \{name.surname\}@stiima.cnr.it}}%
\thanks{$^{2}$ Dipartimento di Ingegneria Meccanica e Industriale, University of Brescia 
{\tt\small \{c.tonola001, manuel.beschi\}@unibs.it}}%
\thanks{$^{3}$Department of Robotics, University of Michigan, Ann Arbor, MI 48109, United States.
{\tt\small mfaroni@umich.edu}}
\vspace{0.2cm} 
}

\maketitle


\begin{abstract}
With the spread of robots in unstructured, dynamic environments, the topic of path replanning has gained importance in the robotics community.
Although the number of replanning strategies has significantly increased, there is a lack of agreed-upon libraries and tools, making the use, development, and benchmarking of new algorithms arduous. This paper introduces {\libname}, a new open-source ROS-based C++ library for sampling-based path replanning algorithms. The library builds a framework that allows for continuous replanning and collision checking of the traversed path during the execution of the robot trajectory. Users can solve replanning tasks exploiting the already available algorithms and can easily integrate new ones, leveraging the library to manage the entire execution.
\end{abstract}


\section{Introduction}
\label{sec: intro}

With the advances in industrial, social, and exploration robotics, robots are increasingly required to work in dynamic environments (e.g., sharing their workspace with humans). 
For this reason, it is necessary for the robot to deploy a reactive behavior, enabling it to react promptly to changes occurring in the work environment. 
For example, a robot sharing the workspace with an operator will often be obstructed by the latter. To avoid collisions and reduce downtime, the robot should be able to promptly change its path without stopping (Fig. \ref{subfig: hrc}).
The replanning problem has been approached in different ways and under different assumptions on the robot and the environment. 
Among the main approaches we can find potential-field, graph-based,  sampling-based,  and reinforcement learning-based methods.
Potential field-based methods use repulsive and pullback forces to deform the robot's trajectory, but can suffer from stagnation in local minima \cite{elastic-band}.
Graph-based methods \cite{LPAStar,D*Lite, AD*} suffer from the curse of dimensionality, while sampling-based methods \cite{MARS,ERRT,DRRT,Anytime_dynamic_RRT,MP-RRT} randomly sample the search space to mitigate this problem. 
Reinforcement learning-based methods learn a policy to achieve the goal while avoiding collisions, but the required training typically does not scale to unknown or unpredictable scenarios \cite{LI2021106446, Nicola-et-al-2021}. 
For these reasons, sampling-based approaches are the most popular methods up to date.

Unlike path planning, path replanning requires real-time updates to the initially calculated plan as the environment continuously changes to ensure safe execution of the robot's trajectory. This is because the path can become invalid during execution in the presence of dynamic obstacles. Thus, a path replanner requires an architecture that continuously monitors the scene, triggers replanning, and contextually executes the robot's trajectory.

There are several libraries dedicated to path planning \cite{OMPL, OPENRAVE, OOPS}, but the same cannot be said for path replanning. 
In particular, OMPL \cite{OMPL} has become the standard library for motion planning.
It implements many state-of-the-art planners and provides integration with other software tools such as \textit{MoveIt!} \cite{moveit}.  
OMPL deals only with static environments; that is, it does not implement any architectures to concurrently handle scene tracking, current path adaptation and trajectory execution, necessary to solve the path replanning problem in dynamic environments.
Recently, \textit{MoveIt!} \cite{moveit} has introduced a hierarchical architecture (\textit{Hybrid Planning}) for adapting a motion plan online. 
It follows a classic hierarchical approach with a global planner that computes an initial trajectory and a local planner that interpolates it and makes local changes. 
When the latter fails, the robot stops and the global planner is queried again.
This hierarchical architecture is a common strategy inspired by mobile robots, yet it prioritizes local changes and invokes the replanner only when the local planner is stuck.
In this paper, we focus on continuous re-planning and execution, for which, to the author's best knowledge, no open-source path replanning libraries are available online.


The goal of the paper is to present {\libname} \cite{replanning_strategies} an open-source C++ library based on ROS \cite{ROS} that implements an entire framework to handle smooth and continuous path replanning, scene tracking and trajectory execution concurrently. 
One can easily use the already available algorithms (the list is expanding) and integrate new ones quickly, without spending time on building the entire necessary architecture. 
The library develops abstract classes that limit the effort required to develop and implement new replanners.

{\libname} can be used to solve replanning problems in dynamic environments, where the initially calculated path can be invalidated by moving obstacles. For example, in Fig. \ref{subfig: hrc} it was used to allow humans and robots to share their workspace in a human-robot cooperation application by modifying online the robot path according to the operator movements \cite{Tonola-roman}. The library was also used in \cite{MARS} to benchmark several replanning algorithms in different scenarios (Fig.\ref{subfig: simulation point}-\ref{subfig: simulation arm}).


\section{Library overview} \label{sec: lib_overview}
The purpose of {\libname} is to provide a tool that makes it easier to use, test, implement, and benchmark sampling-based path replanning algorithms for research and education. 
For this reason, its development took into consideration the following aspects:
\begin{itemize}
    \item \textit{Efficiency}: the library is developed in C++ to obtain a reliable and fast tool, features needed to get quick reactions from the robot;
    \item \textit{Easy to use}: available replanners can be used easily, little code is needed to launch an execution;
    \item \textit{Easy development}: new sampling-based path replanning algorithms can be easily integrated into the framework, using the guidelines and tools provided by the abstract classes;
    \item \textit{Easy integration}: the library is integrated with ROS, so additional software can interact with the framework (\textit{e.g.}, replanning and safety speed monitoring runs together in \cite{Tonola-roman})
\end{itemize}

As mentioned in Section \ref{sec: intro}, the library mainly develops a framework to handle replanning, scene monitoring and trajectory execution simultaneously. 
In addition, the library has features that aid in debugging and benchmarking, such as path visualization, simulation of random obstacles, and collection of useful data. 
{\libname} has a few dependencies, many of which are simple internally implemented ROS packages that support the architecture. 
The main one is \textit{graph\_core}, a library that defines the classes needed for a path planning problem and some sampling-based algorithms to solve it. 
The only external dependencies are Eigen \cite{eigenweb} and \textit{MoveIt} \cite{moveit}, which is used to track the planning scene and for collision checking. 
Although it is well known and used among the ROS community, someone may want to use some other software. Therefore, generalization of this dependency is being evaluated.

\section{Core concepts} \label{sec: core_concepts}

\begin{figure*}[t]
\centering
 \includegraphics[width=0.55\textwidth]{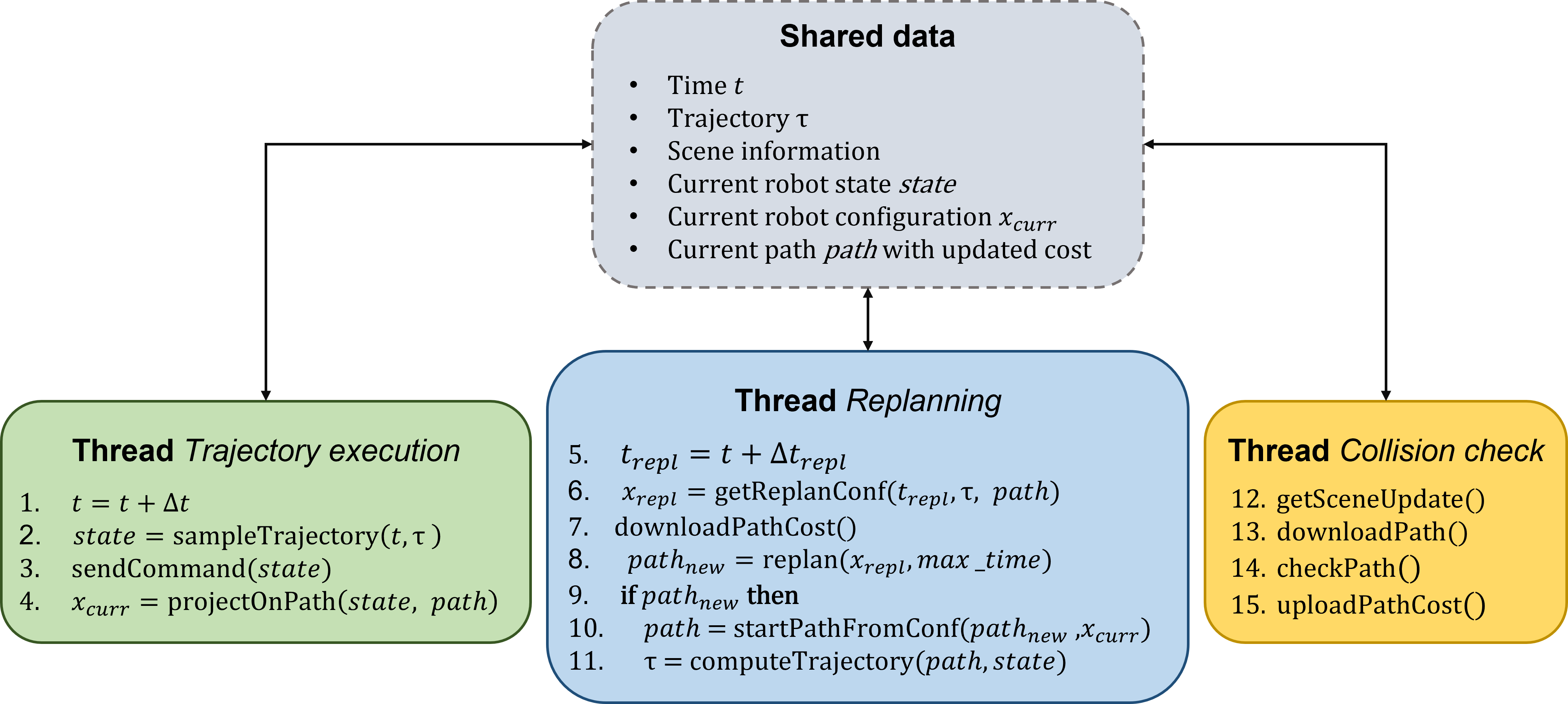}
\caption{Conceptual overview of the \texttt{replanner\_manager}} 
\label{fig: framework}
\vspace{-0.5cm}
\end{figure*}

The two main players in the library are the \texttt{replanner} and \texttt{replanner\_manager}. The \texttt{replanner} provides an abstraction for the path replanning algorithm, and the \texttt{replanner\_manager} defines the entire architecture.

\subsection{The replanner}
The \texttt{replanner} is the actor that executes the replanning algorithm when commanded by the \texttt{replanner\_manager}. 
\texttt{replanner} provides a base class in which to define the replanning algorithm, but its implementation is up to the user. It defines the structure that a replanner must have to interact with the \texttt{replanner\_manager} and requires that the \texttt{replan} function be implemented.
The information the \texttt{replanner} needs are:
\begin{itemize}
    \item the configuration of the robot: a new path will be searched starting from this configuration;
    \item the current path and/or tree: they are usually used to find quickly a new solution by replanning algorithms;
    \item the maximum replanning time: to get responsive behavior from the robot, the algorithm should provide a solution in a short amount of time; 
    \item the path planning problem solver: the algorithm used to find a path from the current configuration to the goal.
\end{itemize}

For example, DRRT \cite{DRRT} is a popular replanning algorithm which deletes the invalid part of the tree and rebuilds it starting from the goal up to the robot configuration. 
It therefore needs to know the tree, the current configuration of the robot, and a solver (in this case RRT \cite{RRT}) to rebuild the tree.

\subsection{The replanner manager}
Fig. \ref{fig: framework} shows a conceptual overview of the architecture of the \texttt{replanner\_manager}, which is composed of three main threads running in parallel:

\begin{itemize}
 \item \textit{Trajectory Execution Thread}: micro-interpolates the robot's current trajectory to send the new command to the robot controller at a high rate.
 \item \textit{Collision Checking Thread}: updates the scene information and checks for the collisions along the path in execution. It considers the part of the current path from the current robot configuration $x_{\mathrm{curr}}$ to the goal.
 \item \textit{Path Replanning Thread}: invokes the \texttt{replanner} to execute the replanning algorithm (line 8) to find a path when the current one is obstructed or to optimize the current solution. If the replanner finds a new path, then it computes a trajectory on it. 
\end{itemize}

These three components  are roughly in charge of execution, collection of obstacle data, and replanning, respectively. 
The \emph{trajectory execution} thread is usually the one that runs the fastest because it is the interface between the framework and the controller. 
Interpolation of the trajectory $\tau$ happens at line 2 of Fig. \ref{fig: framework}. 
Then, the interpolated state is projected on the nominal path (line 4). 
The reason for this is that time parameterization algorithms usually introduce small deviations from the path computed by the path planner (e.g., blending radii), while the replanning algorithm reasons on nodes that strictly belong to the path or the tree. 
In parallel, the \textit{collision check} thread takes a snapshot of the scene (line 12), updates the path cost based on this scene (line 14), and shares the information with the \emph{replanning} thread (line 15), which will rely on this snapshot and the current path cost calculated to search for a new solution. 
Finally, the replanning algorithm defined by the \texttt{replanner} determines how the new path is search and/or improved (line 8). 


Information is exchanged between threads by means of shared data (grey box in Fig. \ref{fig: framework}). Specifically, each thread owns a copy of the current path, which is updated whenever it is updated in the shared data. 
This allows threads (especially the replanning and collision check threads) to carry out their tasks without interfering. 
There is therefore a mechanism for downloading and uploading the shared information at the beginning and at the end of the threads. 
For example, the collision checking thread updates the local copy of the path, if necessary, at the beginning of each iteration (line 13). After that, it uploads the calculated cost and scene information into the shared path with the other threads (line 15). 
The replanning thread therefore downloads this information at the beginning of the iteration (line 7) and at the end uploads the new calculated path and trajectory.

Because the replanner has a planning latency while the robot is moving, it is necessary to consider the displacement between the expected and the real position at the end of the replanner query.
So, to obtain a smoother transition from the current path to the new one, the \texttt{replanner} replans by considering as if the robot were in a state later in time than the current one (line 8). 
To do that, the replanning configuration $x_{\mathrm{repl}}$ is obtained by sampling the trajectory at time instant $t+ \Delta t_{\mathrm{repl}}$ and then projecting the state on the current path. 
This is done by \texttt{getReplanConf} at line 6.  Consequently, when a new path is found, it must be adapted to have the most recent $x_{\mathrm{curr}}$ as starting configuration (\texttt{startPathFromConf} function at line 10). 
$\Delta t_{\mathrm{repl}}$ should be a value equal to or slightly greater than the maximum time given to the replanner to find a solution. The trajectory is then computed considering the robot's current conditions, represented by $state$ at line 11.



\subsection{Development of a new replanner}
The implementation and integration into the framework of a new replanning algorithm can be summarized with the following pipeline:
\begin{enumerate}
    \item implementation of a \texttt{replanner}'s child class and its \texttt{replan} function;
    \item implementation of a  \texttt{replanner\_manager}'s child class that contains and initializes the \texttt{replanner} object;
    \item definition of the triggering condition of the \texttt{replanner} (e.g., replan when there are collisions along the current path or continuously to refine the current solution);
    \item implementation of the \texttt{startPathFromConf} function: it defines how to set $x_{\mathrm{curr}}$ as starting configuration of the path. This could happen simply by adding a node in the tree and extrapolating the path to the goal, but in general it strongly depends on the implemented replanning algorithm, so the definition of the function is left to the user. Note that this function should not be computationally expensive compared to the actual replanning (line 8);
\end{enumerate}

Clearly this represents the minimum requirement to be able to integrate a new algorithm within the framework. However, in a replanning context there are many important aspects, \textit{i.e.} the generation of the trajectory and the management of mobile obstacles for simulations. Default functions are already provided and may be suitable for many cases, but the user has the possibility to customize and override them.

\subsection{Debugging and visualization}
The library has  useful features for debugging and benchmarking. Among these we find the real time visualization of the current path on RViz and the random appearance of unexpected obstacles, as shown in Fig. \ref{subfig: simulation arm} and \ref{subfig: simulation point}. A data collection thread is also available to collect data for benchmarking.
In particular, the latter carries out statistical analyzes on the replanning time during an entire execution, the length of the path actually traversed, the success or failure of the execution and possibly the number of obstacles with which a collision has occurred. All of this, together with different levels of verbosity, can be turned on or off by parameters.


\subsection{Available replanners}
The library currently implements some popular algorithms, such as MARS \cite{MARS}, DRRT \cite{DRRT}, Multi Parallel-RRT \cite{Multi-parallel-RRT}, Anytime DRRT \cite{Anytime_dynamic_RRT}, and \cite{Connell:DRRT*}, which have been successfully used for benchmarking in \cite{MARS}. One can take advantage of the implementation provided in the library to have a starting point for new developments.

\section{Example applications} \label{sec: example_application}
\begin{figure}[t]
\centering
 \includegraphics[width=1\columnwidth]{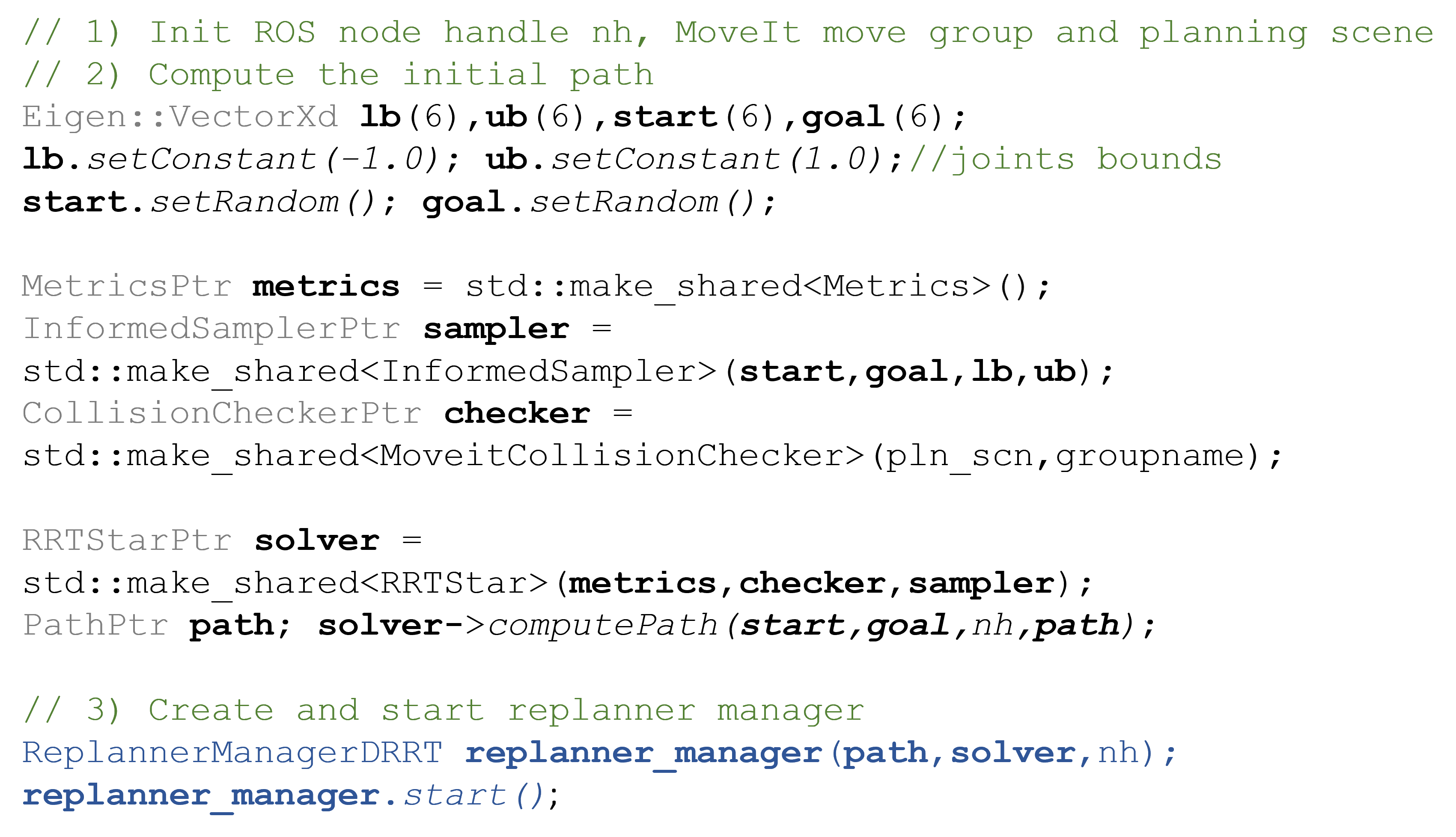}
\caption{Simple C++ code to solve a path replanning problem.} 
\label{fig: code}
\vspace{-0.5cm}
\end{figure}

Fig. \ref{fig: code} shows a simple C++ example to solve a path replanning problem. First, create a ROS node handle and initilize the \textit{MoveIt!} move group. Then, set the robot joints bounds, start and goal configurations, and initialize the metric (which is used to evaluate the path's cost, \textit{i.e.} Euclidean norm), the sampler and the collision checker. Note that the latter needs a planning scene object to evaluate collisions. Now you can use a path planning solver to find the initial path. Finally, create the replanner manager and start the trajectory execution with replanning. More details and tutorials can be found at \cite{replanning_strategies}.

Fig. \ref{subfig: hrc} shows an application of the framework in a human-robot collaboration context. 
A vision system identifies the human's skeleton and publishes bounding spheres to the \textit{MoveIt!} scene. Based on this, the replanner is able to alter the path to avoid the operator. Thanks to the integration with ROS, it was possible to make the replanner interact easily with an ISO/TS15066-based speed scaling module launched on a different node. The replanner was inserted into a controller in the ROS Control framework \cite{ros_control} which read the commands from the \texttt{replanner\_manager} and sent them to the robot. Fig. \ref{subfig: simulation arm} and \ref{subfig: simulation point} are snapshots from a simulated environment, where robots are asked to move from a start configuration to a goal configuration while some unexpected obstacles appear. 

\section{Conclusions and future works} \label{sec: conclusions}
This paper proposes {\libname}, an open-source C++ ROS-based library for easy use and deployment of sampling-based path replanning algorithms. 
The library offers a comprehensive framework for managing trajectory execution with continuous replanning and collision checking of the current path. 
Additionally, it provides a range of valuable tools for algorithm usage, implementation, debugging, and benchmarking. 
The primary objective of this library is to offer researchers and students an off-the-shelf architecture to use and develop online motion planning algorithms.
The library is actively under development and can be accessed at \cite{replanning_strategies}. 
Ongoing efforts involve the creation of extensive documentation and tutorials to enhance user-friendliness. 
Further developments under evaluation are the generalization of the software used for scene monitoring and the integration with OMPL. A ROS-free version of the library is being developed to allow integration with other frameworks and facilitate migration to ROS2, offering users platform flexibility.

\bibliographystyle{IEEEtran}
\bibliography{reference_short}

\end{document}